\documentclass[lettersize,10pt,journal,twoside]{IEEEtran}
\usepackage{ieee_robotics}
\usepackage{bbm}

\usepackage{orcidlink}
\acrodef{rl}[RL]{Reinforcement Learning}
\acrodef{dof}[DoF]{Degrees of Freedom}
\acrodef{moe}[MoE]{Mixture of Experts}
\newcommand{\goalpose}{\mathbf{g}}
\newcommand{\goalpos}{\goalpose^{P}_{pos}}
\newcommand{\goalori}{\goalpose^{P}_{ori}}
\newcommand{\goalcontact}{\goalpose_{fj}}

\newcommand{\thres}{\mathbf{\epsilon}}
\newcommand{\posthres}{\thres_{pos}}
\newcommand{\orithres}{\thres_{ori}}
\newcommand{\contactthres}{\thres_{fj}}

\newcommand{\indicator}{\mathbbm{1}}

\newcommand{\joint}{\mathbf{J}}
\newcommand{\armjoint}{\joint^{a}}
\newcommand{\handjoint}{\joint^{h}}
\newcommand{\base}{\mathbf{b}}
\newcommand{\pbase}{\base_{p}}
\newcommand{\qbase}{\base_{q}}
\newcommand{\obj}{O}
\newcommand{\obs}{o}
\newcommand{\objpose}{\obs^{P}}
\newcommand{\objpos}{\objpose_{p}}
\newcommand{\objori}{\objpose_{q}}
\newcommand{\simstep}{s_i}

\newcommand{\distance}{\phi}
\newcommand{\posdist}{\distance_{p}}
\newcommand{\oridist}{\distance_{\theta}}
\newcommand{\contactdist}{\distance_{j}}

\newcommand{\rew}{r}
\newcommand{\normrew}{\tilde{\rew}}
\newcommand{\distrew}{\rew_{\text{dist}}}
\newcommand{\actionpen}{\rew_{\text{ap}}}
\newcommand{\succrew}{\rew_{\text{succ}}}
\newcommand{\scale}{w}
\newcommand{\posrewscale}{\scale_p}
\newcommand{\orirewscale}{\scale_\theta}
\newcommand{\contactrewscale}{\scale_j}
\newcommand{\actionpenscale}{\scale_{ap}}
\newcommand{\succrewscale}{\scale_{succ}}

\newcommand{\traj}{\tau}
\newcommand{\horizon}{\mathbf{T}}
\newcommand{\predictionhorizon}{\horizon_p}
\newcommand{\obshorizon}{\horizon_o}
\newcommand{\actionhorizon}{\horizon_a}
\newcommand{\stepsize}{\mathbf{L}}
\newcommand{\sequence}{\mathbf{S}}
\newcommand{\timestep}{\mathbf{t}}
\newcommand{\noise}{\mathbf{n}}
\newcommand{\diffusionobs}{\obs^{D}}
\newcommand{\noisenet}{\noise_\theta}

\newcommand{\aset}{\mathcal{A}}

\newcommand{\E}{\mathbb{E}}

\newcommand{\policyaction}{\mathbf{a}}
\newcommand{\baseaction}{\policyaction^{b}}
\newcommand{\handaction}{\policyaction^{h}}

\title{\LARGE \bf Dexterous Functional Pre-Grasp Manipulation with Diffusion Policy}

\begin{document}

\author{Tianhao Wu\orcidlink{0009-0007-9353-4093}, Yunchong Gan\orcidlink{0009-0001-0467-0904}, Mingdong Wu\orcidlink{0009-0007-9120-4621}, Jingbo Cheng\orcidlink{0009-0007-6519-0510}, Yaodong Yang\orcidlink{0000-0001-8132-5613}, Yixin Zhu\orcidlink{0000-0001-7024-1545}, and Hao Dong\orcidlink{0000-0002-7984-9909}
\thanks{Manuscript received xxx April 2024; accepted xx xxxxxx 2024. Date of publication xx xxxxxx 2024; date of current version xx xxxxxx 2024. This work was supported in part by the National Natural Science Foundation of China  (62376006) and the National Youth Talent Support Program (8200800081). T. Wu and Y. Gan contributed equally. Corresponding author: \textit{Hao Dong}.)}%
\thanks{Tianhao Wu, Mingdong Wu, and Hao Dong (e-mail: hao.dong@pku.edu.cn) are with the Center on Frontiers of Computing Studies, School of Computer Science, Peking University, Beijing 100871, China, also with PKU-Agibot Lab, School of Computer Science, Peking University, Beijing 100871, China, and also with National Key Laboratory for Multimedia Information Processing, School of Computer Science, Peking University, Beijing 100871, China.}
\thanks{Yunchong Gan and Jingbo Cheng are with the Center on Frontiers of Computing Studies, School of Computer Science, Peking University, Beijing 100871, China.}
\thanks{Yaodong Yang and Yixin Zhu are with the Institute for Artificial Intelligence, Peking University, Beijing 100871, China.}%
\thanks{See additional material on https://unidexfpm.github.io}%
\thanks{This letter has supplementary downloadable material available at https://doi.org/10.1109/LRA.2024.xxxxxx, provided by the authors.}
\thanks{Digital Object Identifier 10.1109/LRA.2024.xxxxxx}%
}%

\markboth{IEEE ROBOTICS AND AUTOMATION LETTERS, VOL. x, NO. x, xxx 2024}{Wu \MakeLowercase{\etal}: Dexterous Functional Pre-Grasp Manipulation with Diffusion Policy}

\maketitle
\thispagestyle{empty}
\pagestyle{empty}

\begin{abstract}
In real-world scenarios, objects often require repositioning and reorientation before they can be grasped, a process known as pre-grasp manipulation. Learning universal dexterous functional pre-grasp manipulation requires precise control over the relative position, orientation, and contact between the hand and object while generalizing to diverse dynamic scenarios with varying objects and goal poses. To address this challenge, we propose a teacher-student learning approach that utilizes a novel mutual reward, incentivizing agents to optimize three key criteria jointly. Additionally, we introduce a pipeline that employs a mixture-of-experts strategy to learn diverse manipulation policies, followed by a diffusion policy to capture complex action distributions from these experts. Our method achieves a success rate of 72.6\% across more than 30 object categories by leveraging extrinsic dexterity and adjusting from feedback.
\end{abstract}

\begin{figure*}[t!]
    \centering
    \includegraphics[width=\linewidth]{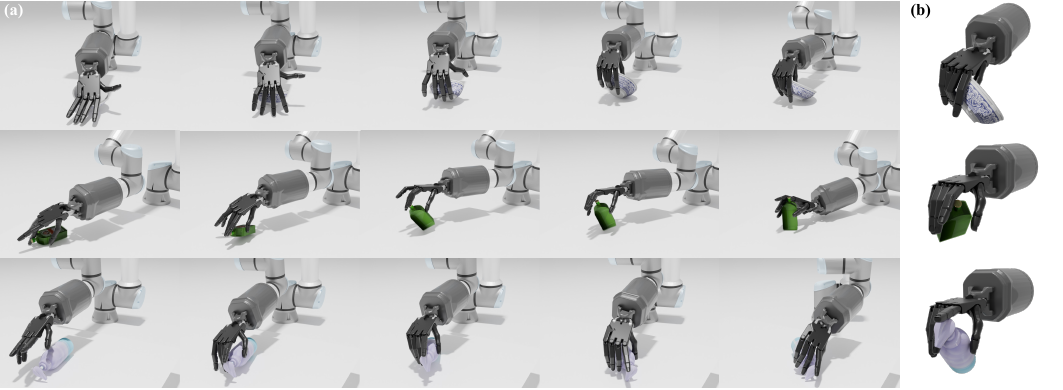}
    \caption{\textbf{Our closed-loop manipulation policy continuously repositions and reorients diverse objects to match the functional grasp goal poses successfully.} (a) The dexterous functional pre-grasp manipulation. (b) The functional grasp goal poses.}
    \label{fig:teaser}
\end{figure*}

\section{Introduction}

Objects in human daily life serve various functions, requiring different functional grasp poses. For instance, when using a spray bottle, one typically positions fingers on the trigger, whereas when passing the bottle to another person, one typically grasps the body. Current works~\cite{wei2023generalized, zhu2023toward} mainly focus on training models to predict the functional grasp pose or further incorporate \ac{rl} for grasp execution and post-grasp usage~\cite{agarwal2023dexterous}. However, these works assume objects are already in highly graspable poses, overlooking the fact that objects are often not positioned with high functional graspability in the real world. For instance, a spray bottle might be lying flat on a table, making it challenging to grasp directly for its intended use. Humans typically manipulate the object into a pre-grasp pose through continuous reorientation and repositioning, a process known as pre-grasp manipulation~\cite{zhou2023learning,chen2023synthesizing}. Unlike conventional pre-grasp manipulation, transitioning objects from ungraspable to graspable states, dexterous functional pre-grasp manipulation further requires both the dexterous hand and the object to satisfy a specific goal pose for subsequent functional grasping.

Dexterous functional pre-grasp manipulation of diverse objects involves intricate interactions with objects and environments, demanding closed-loop dexterous manipulation skills. Existing methods~\cite{huang2023dynamic,chen2022system,chen2023visual} rely on \ac{rl} to train policies for general dexterous manipulation, typically focusing on satisfying the goal orientation and/or position of the objects. However, for functional use, goals must precisely align with the relative position, orientation, and contact between the dexterous hand and the object. This results in an exceedingly small solution space, making it challenging for \ac{rl} agents to explore successful policies. In this scenario, conventional approaches, such as adding distance rewards~\cite{chen2022system,qin2023dexpoint,bao2023dexart}, struggle. Simply adding multiple distance rewards often makes \ac{rl} agents trapped in local minima, failing to devise manipulation policies that meet all criteria. It is also impossible to design specific rewards according to each object~\cite{chen2022towards}, since we need to generalize to diverse objects with diverse poses. Such generalization is also challenging for \ac{rl} agents to learn from scratch~\cite{wu2023learning,xu2023unidexgrasp,wan2023unidexgrasp++}.

To address the problem, we propose a novel mutual reward that computes a scale according to the distance of each criterion and uses the lowest scale to restrict all distance rewards, preventing the agent from getting stuck at a local minimum. Moreover, to facilitate generalization across diverse objects and functional grasp poses, we employ the teacher-student learning framework~\cite{chen2022system,wan2023unidexgrasp++} by training \ac{moe}. The \ac{moe} generates diverse manipulation behavior, leading to a complex action distribution, especially for a dexterous hand with high \acp{dof}. Thus, we propose using a diffusion policy~\cite{chi2023diffusionpolicy}, which has shown great generative modeling ability to capture such complex action distributions.

Through mutual reward and mixture-of-experts training, we observe significant improvements in teacher policy learning. When distilling the teacher policy into a single-student policy using a diffusion policy, our approach achieves teacher-level performance even without object geometry. Our learned policy demonstrates adept use of extrinsic dexterity, such as leveraging tables and inertia to manipulate objects effectively, and also learns to adjust from feedback. These capabilities enhance the policy's ability to generalize across diverse objects.

In summary, our contributions are as follows: (i) We propose a novel mutual reward to address the local minimum problem, significantly improving teacher policy learning. (ii) We propose a pipeline integrating \ac{moe} and diffusion policies to learn complex and general dexterous manipulation policies. (iii) We achieve a general dexterous functional pre-grasp manipulation policy with a 72.6\% success rate across 30+ object categories encompassing 1400+ objects and 10k+ goal poses.

\section{Related Work}

\subsection{Dexterous Functional Grasping}

Dexterous functional grasping is crucial for humans due to the diverse functionalities of objects in real-world scenarios. This encompasses both functional grasp pose generation and execution. Since the functionality of the objects is related to human design, recent frameworks~\cite{zhu2021toward,zhu2023toward} synthesize functional grasp poses using human-labeled part-level functional information. High-quality functional grasping datasets leverage human priors~\cite{wei2023generalized,hang2024dexfuncgrasp} have also been introduced for learning these poses. Additionally, functional affordance regions can be predicted using human functional grasping dataset~\cite{mandikal2021learning,mandikal2022dexvip} or internet data~\cite{agarwal2023dexterous} to indicate functionality.

To address execution, current works mainly rely on \ac{rl} to learn a closed-loop policy by matching functional grasp regions~\cite{mandikal2021learning,mandikal2022dexvip} or setting to a pre-grasp pose~\cite{agarwal2023dexterous} according to functional grasp region. However, these works often assume objects are already positioned for easy functional grasping, ignoring the fact that objects are often not positioned with high functional graspability in the real world, neglecting the need for complex dexterous pre-grasp manipulation.

Our work focuses on dexterous functional pre-grasp manipulation, complementing existing works and serving as a foundation for achieving functional grasping in the real world.

\subsection{Dexterous Manipulation}

Dexterous manipulation presents a significant challenge due to the need for closed-loop policies for handling complex and discontinuous contacts, which are notoriously difficult to model accurately. Model-free \ac{rl} has emerged as a popular approach for acquiring dexterous manipulation skills, as it bypasses the need for explicit contact modeling~\cite{rajeswaran2017learning,jain2019learning,qin2022dexmv,chen2022system,hu2023dexterous,chen2023sequential,yin2023rotating}.

This approach has demonstrated generalization across diverse objects and goals by shaping different distance rewards to enhance exploration. An orientation distance reward has been used for learning general in-hand reorientation~\cite{chen2022system,chen2023visual}, while a position distance reward is used for learning general dynamic handover~\cite{huang2023dynamic}. Combining both rewards has achieved general in-hand manipulation of slender cylindrical objects~\cite{hu2023dexterous}. For general articulated object manipulation, the distance reward between the dexterous hand palm and object part has been applied to enhance reaching the goal part~\cite{bao2023dexart}.

However, our task involves manipulating both arm and dexterous hands to achieve precise position, orientation, and contact goals, resulting in a narrow solution space. Conventional distance rewards can easily trap \ac{rl} agents in local minima. Moreover, our work requires generalization to diverse objects and goals, making it difficult to design specific rewards for each object.

\subsection{(Dexterous) Pre-grasp Manipulation}

Pre-grasp manipulation has been extensively researched to enhance graspability by leveraging extrinsic dexterity. Most works focus on designing specific pre-grasp manipulation strategies to improve graspability. For parallel grippers, \ac{rl}-based systems utilize external surfaces like tables and walls to transform ungraspable objects into graspable states~\cite{zhou2023learning}. Support surfaces and secondary arms can also be employed to achieve power grasps for objects on a table~\cite{baek2021pre}. Additionally, obstacles can be adjusted to improve graspability~\cite{moll2017randomized}. 

Pre-grasp manipulation using a dexterous hand can develop more pre-grasp manipulation strategies, such as a push-and-grasp strategy where a dexterous hand pushes an ungraspable object occluded by the environment to a graspable state before grasping~\cite{dogar2010push}. Another framework involves pushing, rotating, and sliding actions tailored to different objects~\cite{kappler2010representation}. For generalization to diverse scenes, a physics-based method has been proposed by leveraging tables and other environmental objects to transform ungraspable objects into graspable ones~\cite{chen2023synthesizing}.

Unlike conventional pre-grasp manipulation, our work focuses on manipulating diverse objects to diverse goal poses for subsequent functional grasping, rather than solely achieving graspability.

\begin{figure*}[t!]
    \centering
    \includegraphics[width=\linewidth]{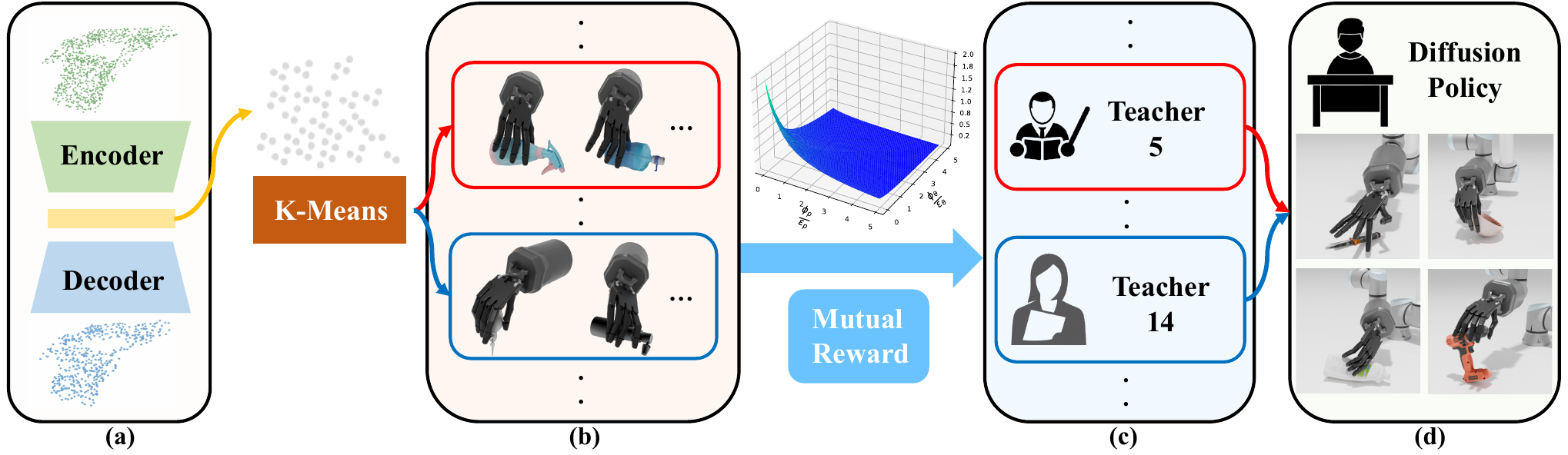}
    \caption{\textbf{Pipeline.} (a) An Autoencoder learns latent representations based on the object-hand point cloud. (b) K-Means clusters the training set into N clusters based on the learned representations. (c) Learning an expert for each cluster based on mutual reward. (d) Distilling multi-expert knowledge into a single student using diffusion for dexterous functional pre-grasp manipulation of seen and unseen objects.}
    \label{fig:pipeline}
\end{figure*}

\subsection{Dexterous Diffusion Models for Grasp and Manipulation}

Diffusion models have demonstrated strong generative modeling capabilities in high-dimensional spaces across various domains~\cite{cai2020learning,song2021solving,yu2022unsupervised,wu2022targf,wang2022humanise,ci2023gfpose,huang2023diffusion,wang2024say}. In dexterous hands, previous studies also show the potential of diffusion models to generate fine-grained, high-dimensional dexterous grasp poses, whether given full~\cite{lu2023ugg,huang2023diffusion} or partial~\cite{weng2024dexdiffuser} object point clouds of a single object. Moreover, diffusion models can handle the complex kinematics and dynamics involved in generating grasp poses for grasping multiple objects with one hand~\cite{li2024grasp}. However, these works mainly focus on pose generation rather than learning manipulation policies.

In closed-loop manipulation policy learning, leveraging the scalability of diffusion models to high-dimensional output space and their ability to express complex action distributions, diffusion policy has been proposed for parallel grippers to acquire dexterous manipulation skills~\cite{chi2023diffusionpolicy, ha2023scaling}. Even for complex, high-\ac{dof} dexterous hand manipulation policies, point cloud-based diffusion policy~\cite{ze20243d,wang2024dexcap} has been introduced, achieving impressive performance. However, these studies are focused on limited objects or a single manipulation goal.

Our focus, however, involves generalization across a wide range of objects and goals, and we leverage diffusion policy for multi-expert teacher-student learning.

\section{Dexterous Functional Pre-Grasp Manipulation}\label{sec:dex-manipulation}

We address the problem of dexterous functional pre-grasp manipulation. Given a functional grasp goal configuration, a policy is tasked to control a robotic arm and dexterous hand to manipulate the object and achieve the specified goal pose.

\paragraph*{State and Action Spaces}

We consider a tabletop scenario with a 6-\ac{dof} robotic arm $\armjoint \in \mathbb{R}^6$ and a 24-\ac{dof} dexterous hand $\handjoint \in \mathbb{R}^{24}$. The hand's base pose is defined as $\base=[\pbase, \qbase]$, where $\pbase \in \mathbb{R}^3$ denotes the 3D position and $\qbase \in \mathbb{R}^4$ the 4D quaternion. The hand joints consist of 2-\ac{dof} wrist joints $\joint^{w} \in \mathbb{R}^{2}$, 18-\ac{dof} finger joints $\joint^{f} \in \mathbb{R}^{18}$, and 4-\ac{dof} underactuated finger joints $\joint^{u} \in \mathbb{R}^{4}$. The action space $\aset \subseteq \mathbb{R}^{26}$ encompasses 6D relative changes for the hand base $\baseaction$ and 20D relative changes for the actuated hand joints $\handaction$.

\paragraph*{Task Simulation}

For each pre-grasp manipulation trial, we sample a desired goal pose $\goalpose$ from a prior goal distribution. Each $\goalpose$ corresponds to a specific object $\obj$, but one $\obj$ can have multiple $\goalpose$. The hand's palm coordinate is denoted as $P$. The goal pose $\goalpose = [\goalpos, \goalori, \goalcontact]$, where $\goalpos \in \mathbb{R}^3$ denotes the relative 3D goal position of the object's center of mass with respect to the hand's palm, $\goalori \in \mathbb{R}^4$ denotes the relative 4D goal quaternion of the object's center of mass with respect to the hand's palm, and $\goalcontact \in \mathbb{R}^{18}$ denotes the angle of the hand's actuated finger joints.

\paragraph*{Observations}

The policy $\pi(\policyaction | \cdot)$ needs to adapt to different $\goalpose$ and $\obj$. Therefore, it conditions on $\armjoint, \base, \handjoint, \goalpose$, and $\objpose = [\objpos, \objori]$, where $\objpos \in \mathbb{R}^3$ denotes the relative 3D position of the object's center of mass with respect to the hand's palm, and $\objori \in \mathbb{R}^4$ denotes the relative 4D quaternion of the object's center of mass with respect to the hand's palm.

\paragraph*{Objective}

The objective of this task is to find a policy $\pi(\policyaction | \base, \armjoint, \handjoint, \objpose, \goalpose)$ that maximizes the expected pre-grasp manipulation success rate:
\begin{equation}
    \pi^* = \arg\max_{\pi}
    \E_{\policyaction_t \sim \pi(\cdot | \base_t, \armjoint_t, \handjoint_t, \objpose_t, \goalpose)}
    \left[ \indicator(\text{success}) \right].
    \label{eq:objective}
\end{equation}
The success is satisfied if $\posdist<=\posthres$ and $\oridist<=\orithres$ and $\contactdist<=\contactthres$, where $\posdist = | \objpos - \goalpos |_2$ is the distance between $\objpos$ and $\goalpos$, $\oridist = 2 \arcsin \left(\left(\objori \cdot (\goalori)^{-1}\right)_4\right)$ denotes the distance between $\objori$ and $\goalori$, and $\contactdist = | \joint^{f} - \goalcontact |_2$ denotes the distance between $\joint^{f}$ and $\goalcontact$. $\posthres$, $\orithres$, and $\contactthres$ denote the distance threshold for position, orientation, and contact.

\section{Method}

In dexterous functional pre-grasp manipulation, the high dimensionality of the dexterous hand leads to a vast policy space. Meanwhile, the task itself presents a limited solution space, as successful manipulation requires achieving precise goals that satisfy position, orientation, and contact criteria.

Despite the success of model-free \ac{rl} in various manipulation tasks~\cite{huang2023dynamic,chen2022system}, the stringent requirements in dexterous functional pre-grasp manipulation pose significant challenges to exploration, especially for agents with limited observations.

To address these challenges, we employ the teacher-student framework~\cite{chen2022system} (see \cref{fig:pipeline}), utilizing a pre-trained ``teacher'' agent with superior knowledge to guide a ``student'' agent during the learning process.

\subsection{Teacher Policy Learning}

Teacher policy learning aims to acquire high-performance experts without restricting access to privileged information~\cite{chen2022system}. We introduce a novel mutual reward for learning dexterous functional pre-grasp manipulation policies, followed by an \ac{moe} to enhance the overall performance of the teacher policy.

\paragraph*{Mutual Reward}

Reward shaping is crucial for training a proficient \ac{rl} agent. In our task, even with privileged information, conventional reward shaping approaches, like adding distance rewards for each goal component~\cite{chen2022system,qin2023dexpoint,bao2023dexart}, can easily trap the \ac{rl} agent in a local minimum. These rewards incentivize the agent to prioritize optimizing easily achievable distance rewards, such as position distance $\posdist$ and contact distance $\contactdist$, by manipulating the hand base and joints. However, the agent tends to neglect the orientation distance $\oridist$, which requires reorienting the object.

To address this, we propose a novel mutual reward. We first define a normalization function $\psi$ to standardize different distance rewards into the range [0,1]:
\begin{equation}
    \normrew = \psi \left( \phi, \thres \right) = \frac{\thres}{ \phi  + \thres},
\end{equation}
where $\normrew$ denotes the normalized distance reward. Given the challenge of defining the optimization order for three distance rewards, we use the minimum normalized distance reward $\normrew_\text{min}$ as a scale to regulate all the distance rewards. Thus, the total distance reward becomes:\begin{equation}
    \distrew = \normrew_\text{min} \left(
        \posrewscale \tilde{r}_p + 
        \orirewscale \tilde{r}_\theta + 
        \contactrewscale \tilde{r}_j
        \right),
\end{equation}
where $\posrewscale$, $\orirewscale$, and $\contactrewscale$ are hyperparameters. This restriction term prevents simply minimizing $\posdist$ or $\contactdist$ from rapidly increasing the total reward, as the typically large $\oridist$ results in a small $\normrew_\text{min}$, as illustrated in \cref{fig:pipeline}. This compels the agent to jointly optimize all three distance rewards, enabling successful learning of the dexterous functional pre-grasp manipulation policy.

We also incorporate an action penalty $\actionpen$ to regulate arm motion:
\begin{equation}
    \actionpen = \| \baseaction \|_2.
\end{equation}
This penalty discourages excessive arm movement and encourages finger utilization for object manipulation. The success reward $\succrew$ is 1 if manipulation is successful. Therefore, the total reward becomes:
\begin{equation}
    \rew = \distrew + \actionpenscale * \actionpen + \succrewscale * \succrew,
\end{equation}
where $\actionpenscale$ and $\succrewscale$ are hyperparameters for the action penalty and success reward, respectively.

\paragraph*{\ac{moe}}

Given the goal of generalizing across diverse objects and goal poses, the manipulation process can exhibit significant diversity. This makes it challenging for \ac{rl} agents to learn a good policy for all goal poses. While Unidexgrasp~\cite{xu2023unidexgrasp,wan2023unidexgrasp++} introduced a framework for learning dexterous grasping for diverse objects by starting with ``GeoCurriculum,'' which gradually increases object instances and categories from a single object with a single pose, such a curriculum is not suitable for our task. Unlike grasping, which involves reaching and closing fingers, manipulation requires continuous repositioning and reorienting of the object. Hence, the manipulation policy for different object geometry can be different. For instance, manipulating a cylindrical bottle involves rolling it, whereas manipulating a camera requires different techniques. Thus, if the agent learns to manipulate a cylindrical bottle first, it may struggle to learn to manipulate the camera.

Although ``GeoCurriculum'' is not directly applicable to our task, the concept of decomposing the task space is valuable. Therefore, we initially cluster the entire task space into several clusters. Unidexgrasp++~\cite{xu2023unidexgrasp} trains an autoencoder on object geometry for the reconstruction task and then uses the latent representation of each object for state-based clustering. In the case of dexterous functional pre-grasp manipulation, the task is linked to the goal pose. Given the same object with the same initial pose, the goal of grasping the handle versus grasping the body can lead to different manipulation processes. Thus, we combine the object and hand point cloud to learn a latent representation.

After clustering, we employ K-Means to partition the entire task space into N clusters. While prior work~\cite{wan2023unidexgrasp++} suggests that a generalist can assist specialists in training dexterous grasping, in dexterous functional pre-grasp manipulation, manipulation behaviors can vary across different goals, such as manipulating a cylindrical bottle versus a camera, as described earlier. Hence, to obtain a specialized high-performance manipulation policy for each cluster, we directly train an expert for each cluster from scratch.

\subsection{Distilling With Diffusion Policy}

Once we have acquired the \ac{moe}, our objective is to distill the diverse manipulation policies into a single student policy. The student policy is constrained to only access observations available in real scenarios, as described in \cref{sec:dex-manipulation}. Given the complexity and diversity of the action distribution resulting from the intricate manipulation process and the \ac{moe}, coupled with the high dimensionality of the dexterous hand, we opt to utilize a diffusion policy~\cite{chi2023diffusionpolicy} which has been shown to have the ability to learn complex high \ac{dof} dexterous hand manipulation behavior~\cite{ze20243d,wang2024dexcap}, to model the complex action distribution of different experts. Diffusion policy formulates the robot behavior generation as a conditional denoising process.

\paragraph*{Dataset Generation}

Since the diffusion policy operates as an offline imitation learning framework, we must gather demonstrations using our teacher experts. While our teacher policy necessitates privileged information for inference, the trajectories we gather for training the diffusion policy solely comprise limited observations.

By executing the policy of our N teacher experts on the entire task space, we sample a set of trajectories ${{ \traj_i }}_{i=1}^M$. However, these trajectories have different episode lengths. Following Chi \etal~\cite{chi2023diffusionpolicy}, for each trajectory $\traj_i$ with a step size of $\stepsize_i$, we sample every sequence with a length of $\predictionhorizon$, where $\predictionhorizon$ denotes the prediction horizon. Consequently, we obtain $\stepsize_i - \predictionhorizon + 1$ trajectory data points from $\traj_i$. By iterating over the trajectory set ${{ \traj_i }}_{i=1}^M$, we can generate the dataset ${{ \sequence_j }}_{j=1}^O$ for diffusion policy training.

\paragraph*{Diffusion Policy Training}

The training process involves sampling data points from the generated dataset. For each sample $\sequence_j$, we randomly sample a time step $\timestep$, and then sample a noise $\noise^{\timestep}$. We consider the first $\obshorizon$ steps of observations from $\sequence_j$ as the observation sequence $\diffusionobs_j$, and take the $\predictionhorizon$ steps of actions from $\sequence_j$ as the action sequence $\mathbf{A}^0_j$. We utilize $\diffusionobs$ as a condition and define the loss function as follows:
\begin{equation}
    \mathcal{L} = \operatorname{MSE}(
        \noise^{\timestep},
        \noisenet(
        \diffusionobs_j, 
        \mathbf{A}^0_j + \noise^{\timestep},
        \timestep
        )
        ),
\end{equation}
where $\noisenet$ is a noise prediction network.

\paragraph*{Action Generation with Diffusion Policy}

Upon training the noise prediction network $\noisenet$, for each simulation step $\simstep$, the DDPM~\cite{ho2020denoising} performs $\timestep$ steps denoising from the noise action sequence $\mathbf{A}^\timestep_{\simstep}$ sampled from Gaussian noise, until obtaining the noise-free action sequence $\mathbf{A}^0_{\simstep}$. Following equation:
\begin{equation}
    \mathbf{A}^{\timestep-1}_{\simstep} = \alpha(
        \mathbf{A}^\timestep_{\simstep} 
        - \gamma\noisenet(\diffusionobs_{\simstep}, \mathbf{A}^\timestep_{\simstep}, \timestep) 
        + \mathcal{N}(0, \sigma^2 I)
    ),
    \label{eq:diffusion-policy-langevin}
\end{equation}
where $\alpha$, $\gamma$, and $\sigma$ are parameters for noise scheduler. We then execute $\actionhorizon$ steps of the denoised action sequence $\mathbf{A}^0_{\simstep}$.

\subsection{Implementation Details} 

\paragraph*{Teacher Policy}

Our \ac{rl} backbone is PPO~\cite{schulman2017proximal}. We configure hyperparameters with $\posrewscale=\orirewscale=\contactrewscale=3$, $\actionpenscale=-0.01$, and $\succrewscale=800$. Privileged information details for teacher policy training are provided in Table~\ref{table:teacher-obs}.

\begin{table}[ht!]
    \centering
    \footnotesize
    \setlength{\tabcolsep}{3pt}
    \caption{\textbf{Teacher Observation.} The superscript $P$ represents the variable is \wrt the hand-palm coordinate.}

\begin{tabular}{@{}cccccc@{}}
\toprule
Variable                            & Dimension & Description                         \\ 
\midrule
$\pbase$                            & (3,)      & hand base positions                 \\
$\qbase$                            & (4,)      & hand base orientations              \\
$\armjoint$                         & (6,)      & arm joint angles                    \\
$\dot{\armjoint}$                   & (6,)      & arm joint velocities                \\
$\handjoint$                        & (24,)     & hand joint angle                    \\
$\dot{\handjoint}$                  & (24,)     & hand joint velocities               \\
${f_p}^{P}$                         & (5, 3)    & fingertip positions (to Palm)       \\
${f_q}^{P}$                         & (5, 4)    & fingertip orientations (to Palm)    \\
$\boldsymbol{v}_{f}$                & (5, 3)    & fingertip linear velocities         \\ 
$\boldsymbol{w}_{f}$                & (5, 3)    & fingertip angular velocities        \\
\midrule
$\obs_p$                            & (3,)      & object position                     \\
$\obs_q$                            & (4,)      & object orientation                  \\
$\objpos$                           & (3,)      & object position (to Palm)           \\
$\objori$                           & (4,)      & object orientation (to Palm)        \\
$\boldsymbol{v}_{\text{o}}$         & (3,)      & object linear velocity              \\
$\boldsymbol{w}_{\text{o}}$         & (3,)      & object angular velocity             \\
$\boldsymbol{bbox}_{\text{object}}$ & (2, 3)    & object boundingbox                  \\ 
\midrule
$\goalpos$                          & (3,)      & target object position (to Palm)    \\
$\posdist$                          & (3,)      & position distance                   \\
$\goalori$                          & (4,)      & target object orientation (to Palm) \\
$\oridist$                          & (4,)      & orientation distance                \\
$\goalcontact$                      & (18,)     & target hand joint angles            \\ 
$\contactdist$                      & (18,)     & joint distance                      \\
\bottomrule
\end{tabular}
    \label{table:teacher-obs}
\end{table}

\paragraph*{\ac{moe}}

For each goal pose $\goalpose_k$ in the set ${ \goalpose_k }_{k=1}^O$, we sample 512 points from the corresponding object mesh and 512 points from the corresponding hand mesh. These point clouds are concatenated and then encoded using PointNet++~\cite{qi2017pointnet++}, and the reconstruction loss is computed with Chamfer Distance. The entire task space is divided into 20 clusters. For each expert, we train on 12000 parallel environments until convergence.

\paragraph*{Diffusion Policy}

We configure $\predictionhorizon=4$, $\obshorizon=2$, and $\actionhorizon=1$, while keeping other parameters the same as~\cite{chi2023diffusionpolicy}. Because we use the relative action for policy learning, we use the transformer backbone~\cite{vaswani2017attention} for handling quick and sharp changes in action sequence~\cite{chi2023diffusionpolicy}.

\section{Experimental Setups}

\subsection{Task Simulation}

\paragraph*{Environment Setup}

We created a simulation environment based on Isaac Gym~\cite{makoviychuk2021isaac} using ShadowHand and UR10e robots. Each environment consists of an object randomly placed on a table, with its mass randomized from 0.01kg to 0.5kg due to the diversity of object categories. A UR10e robot is positioned outside the table with the ShadowHand mounted on the end of the arm, as shown in \cref{fig:teaser}. The maximum episode length is 300 steps. Episodes terminate upon reaching the goal pose or prematurely if the object falls off the table or the maximum steps are reached.

\paragraph*{Goal Pose Generation}

We utilize the Oakink dataset~\cite{yang2022oakink}, which covers diverse functional intents for a wide range of objects, to generate a dexterous hand functional grasp pose dataset. Since the Oakink dataset is based on the human hand, it differs in structural and shape characteristics from robotic hands. To adapt the hand poses, we employ a retargeting algorithm~\cite{qin2022dexmv} based on task space vectors to map the mano hand pose to the ShadowHand pose. Next, to refine poses prone to collision and non-force closure grasp, we utilize Dexgraspnet~\cite{wang2023dexgraspnet} for optimization. To enhance grasping, we optimize the joints by calculating the gradient corresponding to the movement of the contact point along the normal direction. Finally, all refined poses undergo validation in a simulated environment to eliminate those unstable under the influence of gravity.

Due to the uneven distribution of object instances within each category in the Oakink dataset, we implement a stratified splitting approach for training and testing sets. Categories with a larger number of instances are randomly divided into a 70\% training set and a 30\% testing set, while categories with fewer instances are split evenly into 50\% for training and testing. Overall, our training set comprises 1026 object instances with a total of 6968 goal poses, while the testing set consists of 443 object instances with a total of 3034 goal poses.

\subsection{Baselines and Metrics}

For the teacher policy, we compare our method with the following:
(i) \textbf{PPO-Sum}: This baseline adopts a sum reward approach, combining three distance rewards for \ac{rl} training, while keeping other rewards the same as \textit{Ours}.
(ii) \textbf{Ours-SE}: Based on our proposed reward, we train a single expert for the entire training set.
(iii) \textbf{Ours-MoEF}: This comparison utilizes a \ac{moe}, but instead of training them from scratch, we fine-tune them from the \textit{Ours-SE} model. Due to computational cost, this comparison is conducted on a subset of our training data.

For comparison based on student observations, we evaluate our method with:
(i) \textbf{PPO-OS} employs PPO as a one-stage method, using the same mutual reward as \textit{Ours} but without teacher-student learning.
(ii) \textbf{Behavior Cloning (BC)} serves as an offline imitation learning framework, learning directly from expert demonstrations via supervised learning.
(iii) \textbf{Dagger}~\cite{ross2011reduction} is an online imitation learning framework that tackles the covariate shift problem through iterative sampling with a learned policy via online interaction.

We employ the success rate as the metric for all comparisons. Our task employs stringent criteria, with $\posthres=1cm$, $\orithres=0.1 rad$, and $\contactthres=0.2rad$, which are challenging thresholds to meet.

\section{Results}

We conducted extensive experiments to validate our approach. \Cref{exp:reward-comparison} compares our proposed reward with a baseline using privileged information. Building upon our reward, \cref{exp:challenge} explores the challenges of learning general dexterous functional pre-grasp manipulation without a teacher-student framework. Within this framework, \cref{exp:teacher,exp:student} evaluate the effectiveness of using \ac{moe} for teacher policy training and a diffusion policy for distillation, respectively. To assess our reliance on object pose observation, \cref{exp:ablation:geo,exp:ablation:robust} present results concerning different geometry types and robustness. Finally, \cref{exp:category} analyzes performance across object categories.

\subsection{Reward Comparison}\label{exp:reward-comparison}

As shown in \cref{table:teacher}, without a mutual reward, the \textit{PPO-Sum} agent fails to learn a successful manipulation policy. We observed that the agent quickly learns to align positions and contacts but is stuck at a local minimum, failing to align orientations. In contrast, \textit{Ours-SE} with a mutual reward prevents premature optimization, encouraging simultaneous optimization of each reward. This leads to a significant improvement in the success rate from 0.0\% to 58.0\%. 

\begin{table}[b!]
    \centering
    \footnotesize
    \setlength{\tabcolsep}{3pt}
    \caption{\textbf{Success Rate of Teacher Policy.} ``All'': trained on the entire training set; ``Sub'': trained on a subset of the training set; ``SE'': single expert; ``\ac{moe}'': mixture of experts; ``Succ (Train)'': success rate on the training set.}
\begin{tabular}{llllr}
\toprule
\textbf{Method}                 & \textbf{Training set} & \textbf{Reward}      & \textbf{Teacher} & \textbf{Succ (Train)} \\ 
\midrule
\textbf{PPO-Sum}                & All                   & Sum                  & SE               &  0.0\%                \\
\textbf{Ours-SE}                & All                   & Mutual               & SE               & 58.0\%                \\
\textbf{Ours-SE (sub)}          & Sub                   & Mutual               & SE               & 55.2\%                \\
\textbf{Ours-MoEF (sub)}        & Sub                   & Mutual               & MoE              & 63.9\%                \\
\textbf{Ours (sub)}             & Sub                   & Mutual               & MoE              & 67.4\%                \\
\rowcolor{gray!20}\textbf{Ours} & All                   & Mutual               & MoE              & 75.0\%                \\ 
\bottomrule
\end{tabular}
    \label{table:teacher}
\end{table}

\subsection{Challenges in Functional Pre-grasp Manipulation}\label{exp:challenge}

To demonstrate the difficulty of learning general dexterous functional pre-grasp manipulation, we conducted experiments using \textit{PPO-OS} based on student observation, incorporating our mutual reward. We trained PPO across varying numbers of objects, with each PPO model trained until convergence or reaching the maximum interaction steps (5.76 billion).

As depicted in \cref{table:obj-abl}, when trained on a single object, the \ac{rl} agent rapidly learns a policy with a nearly 100\% success rate. However, as the number of objects increases, the success rate declines steeply, highlighting the difficulty of \textbf{general dexterous functional pre-grasp manipulation}. Interestingly, the success rate for 9 objects is lower than for 100 objects. This is because within the set of 9 objects, the presence of challenging objects, such as knives, is proportionately higher, hindering exploration. This underscores the necessity of employing an \ac{moe}.

\begin{table}[t!]
    \centering
    \footnotesize
    \setlength{\tabcolsep}{3pt}
    \caption{\textbf{Success Rate of One-stage PPO under Different Sizes of Training Set.} ``Succ (Train)'': success rate on the training set. As the number of objects increases, finding a general manipulation policy across diverse objects becomes increasingly challenging for one-stage PPO.}
    \begin{tabular}{lllll}
\toprule
\textbf{Obj Num}      & \textbf{1} & \textbf{9} & \textbf{100} & \textbf{1026 (All)} \\ 
\midrule
\textbf{Succ (Train)} & 98.2\%     & 6.2\%      & 21.2\%       & 6.5\%               \\ 
\bottomrule
\end{tabular}
    \label{table:obj-abl}
\end{table}

\subsection{Teacher Policy Comparison}\label{exp:teacher}

\Cref{table:teacher} demonstrates that employing multiple experts leads to a further increase in the success rate from 58.0\% to 75.0\% compared to a single expert. This is because multiple experts allow the agent to learn more tailored policies for each cluster. Additionally, our experiment shows that training from scratch outperforms fine-tuning from a generalist. We sampled five clusters with varying learning difficulty and trained \textit{Ours (sub)} from scratch on each cluster, while \textit{Ours-MoEF (sub)} was fine-tuned from the pre-trained single expert \textit{Ours-SE (sub)}. As shown in \cref{table:teacher}, training from scratch achieves better overall performance due to the diversity of objects and poses and the complexity of manipulation, making it challenging to transfer a general policy to objects and poses with significant variability.

\subsection{Student Policy Comparison}\label{exp:student}

All methods utilizing teacher-student learning outperform \textit{PPO-OS}, which undergoes end-to-end training. As indicated in \cref{table:main}, Dagger can achieve performance comparable to the single-expert teacher policy but struggles to learn an effective policy under the mixture of expert teacher policy.

\begin{table}[b!]
    \centering
    \footnotesize
    \setlength{\tabcolsep}{3pt}
    \caption{\textbf{Success Rate of Student Policy.} This table presents the results of methods that require teachers for training. ``SE'': single expert; ``\ac{moe}'': mixture of experts;  ``Demo Num'': the maximum number of demonstrations collected for each pose in the training set, used for distilling the student policy; ``Succ (Train)'': success rate on the training set; ``Succ (Test)'': success rate on the testing set.}
        \begin{tabular}{llrrr}
\toprule
\textbf{Method}                  & \textbf{Teacher} & \textbf{Demo Num} & \textbf{Succ (Train)} & \textbf{Succ (Test)} \\
\midrule
\textbf{Dagger}                  &  SE              & -                 & 52.2\%                & 52.3\%               \\
\textbf{Dagger}                  & MoE              & -                 & 17.5\%                & 17.3\%               \\
\textbf{BC}                      & MoE              &  5                & 57.4\%                & 54.7\%               \\
\textbf{BC}                      & MoE              & 10                & 67.7\%                & 65.1\%               \\
\textbf{BC}                      & MoE              & 20                & 70.9\%                & 67.7\%               \\
\textbf{Ours}                    & MoE              &  5                & 65.7\%                & 63.3\%               \\
\textbf{Ours}                    & MoE              & 10                & 71.3\%                & 68.4\%               \\
\rowcolor{gray!20}\textbf{Ours}  & MoE              & 20                & 73.7\%                & 70.1\%               \\
\bottomrule
\end{tabular}

    \label{table:main}
\end{table}

Offline imitation learning methods demonstrate superior results compared to those requiring environment interaction. Due to the critical role of data quantity, we compare \textit{Ours} and \textit{BC} across various demonstration numbers. \Cref{table:main} shows that \textit{Ours} consistently outperforms \textit{BC} on both training and testing sets, especially with limited demonstrations. Notably, \textit{Ours} can achieve comparable performance with only half the number of demonstrations required by \textit{BC}. With a large number of demonstrations, \textit{Ours} approaches teacher-level performance. 

\subsection{Ablation on Geometry Type}\label{exp:ablation:geo}

\begin{figure}[t!]
    \centering
    \includegraphics[width=\linewidth]{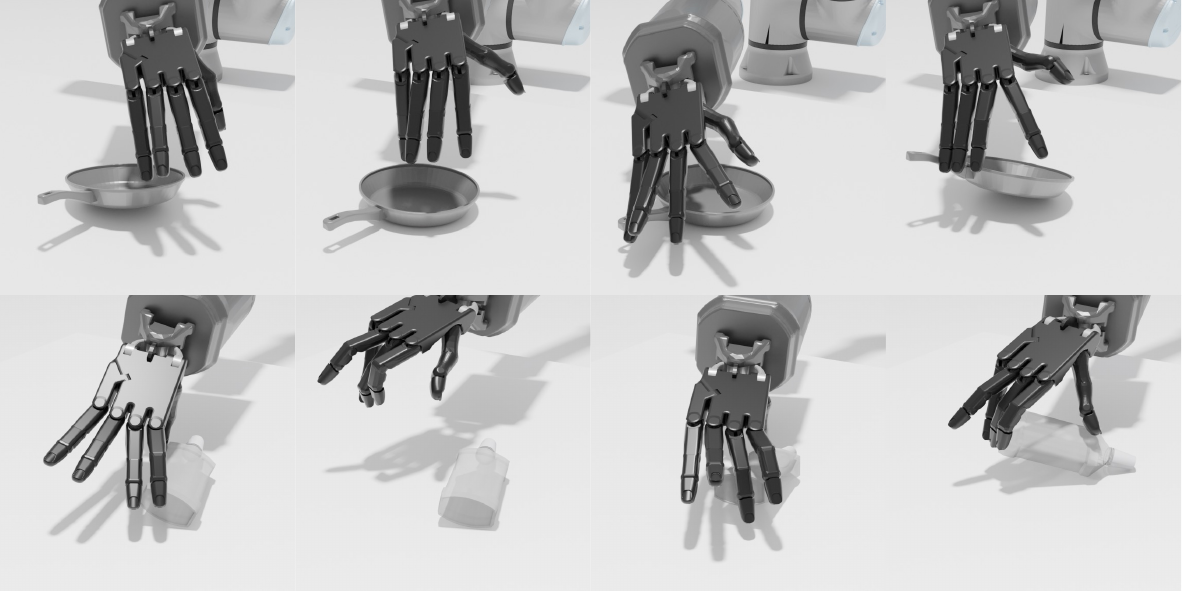}
    \caption{\textbf{Adaptability of Our Learned Policy.} Although our agent may initially fail to manipulate objects, it adjusts its policy on the second attempt, successfully manipulating them. This capability helps the agent handle diverse dynamics.}
    \label{fig:adjust}
\end{figure}

While common sense suggests that object geometry information is crucial for manipulation, \cref{table:geo} shows that providing more detailed geometry information does not significantly impact performance, although it can lead to a better student policy. Observing the learned policy, we discovered that our policy utilizes extrinsic dexterity, such as using the table to roll objects or leveraging inertia, as shown in \cref{fig:teaser}. Additionally, our policy learns to adjust based on feedback (\cref{fig:adjust}). These capabilities enhance the agent's ability to generalize to different objects and goal poses.

\begin{table}[b!]
    \centering
    \footnotesize
    \setlength{\tabcolsep}{3pt}
    \begin{minipage}{.5\linewidth}
        \caption{\textbf{Success Rate of Different Geometries.} ``Succ (Train)'': success rate on the training set; ``Succ (Test)'': success rate on the testing set. Due to computational cost, this experiment was conducted using 5 demonstrations per pose.}
        \resizebox{\linewidth}{!}{
            \begin{tabular}{lcc}
\toprule
\textbf{Geometry Type}       & \multicolumn{1}{l}{\textbf{Succ (Train)}} & \multicolumn{1}{l}{\textbf{Succ (Test)}} \\ 
\midrule
\textbf{Pose + Point Cloud}  & 66.5\%                                    & 63.3\%                                   \\
\textbf{Pose + Bounding Box} & 65.9\%                                    & 62.8\%                                   \\
\textbf{Pose}                & 65.7\%                                    & 63.3\%                                   \\ 
\bottomrule
\end{tabular}%
        }%
        \label{table:geo}
    \end{minipage}%
    \hfill
    \begin{minipage}{.45\linewidth}
        \caption{\textbf{Success Rate under Different Levels of Object Pose Estimation Noise and Success Threshold.} ``Succ (Test)'': success rate on the testing set. The noise level is determined by the standard deviation of the specified noise.}
        \resizebox{\linewidth}{!}{
            \begin{tabular}{crrr}
\toprule
\multicolumn{1}{c}{\multirow{6}{*}{\textbf{Noise level}}} & \multicolumn{3}{c}{\textbf{Threshold}}                   \\ 
\cmidrule{2-4} 
\multicolumn{1}{c}{}                                      & {\begin{minipage}[b]{0.1\columnwidth}\centering {\includegraphics[width=0.9\textwidth]{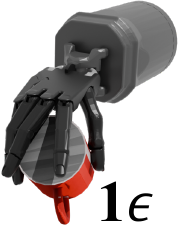}}\end{minipage}}                & {\begin{minipage}[b]{0.1\columnwidth}\centering {\includegraphics[width=0.9\textwidth]{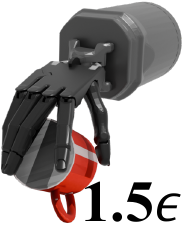}}\end{minipage}}              & {\begin{minipage}[b]{0.1\columnwidth}\centering {\includegraphics[width=0.9\textwidth]{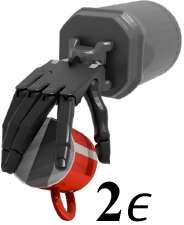}}\end{minipage}}                                                                           \\ 
\cmidrule{2-4} 
\multicolumn{1}{c}{}                                      & \multicolumn{3}{c}{\textbf{Succ (Test)}}                \\ 
\midrule
\textbf{0$^{\circ}$, 0cm}                                 & 70.1\%           & 77.8\%             & 81.2\%           \\
\textbf{2$^{\circ}$, 2cm}                                 & 38.1\%           & 67.7\%             & 75.2\%           \\
\textbf{5$^{\circ}$, 5cm}                                 &  0.0\%           &  6.5\%             &  8.7\%          \\ 
\bottomrule
\end{tabular}

%
        }%
        \label{table:robust}
    \end{minipage}%
\end{table}

However, these capabilities also have drawbacks. We observed instances where the agent pushes objects down to better utilize extrinsic dexterity, which may need improvement in the future through the design of new reward mechanisms.

\subsection{Robustness under Noisy Object Pose}\label{exp:ablation:robust}

As we solely depend on object pose for dexterous functional pre-grasp manipulation, and object pose is actually hard to be accurate in the real world due to sensor noise and occlusion, we further conduct experiments under varying levels of noisy object pose observations~\cite{chen2022epro}. 

\Cref{table:robust} shows that injecting 2$^{\circ}$, 2cm noise results in a decrease in success rate. However, given our stringent criteria, we also tested with a larger success threshold, which remains reasonable. Even when doubled, the achieved functional pose remains meaningful and comparable. By slightly relaxing the threshold, our method still achieves a high success rate, underscoring its robustness and potential for real-world applications.

\subsection{Performance across Object Categories}\label{exp:category}

\begin{figure}[t!]
    \centering
    \includegraphics[width=\linewidth]{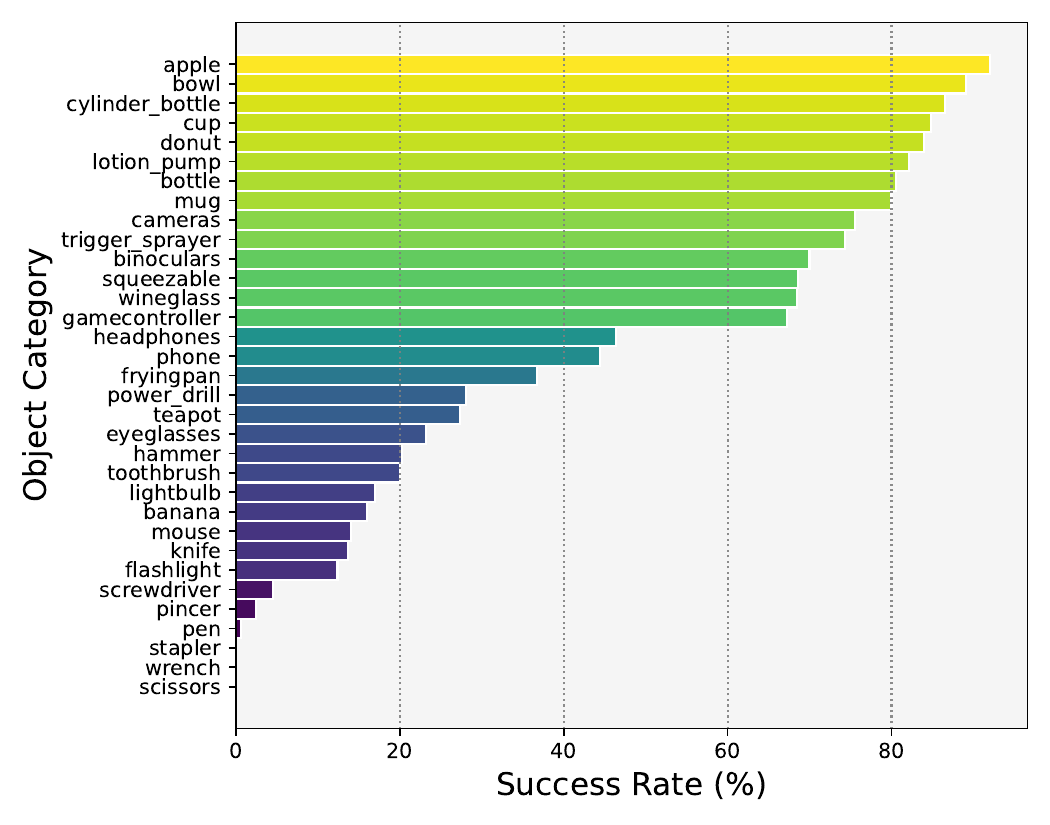}
    \caption{\textbf{Success Rate of Different Object Categories.} 
    }
    \label{fig:succ-per-cat}
\end{figure}

\Cref{fig:succ-per-cat} shows that while our method achieves a high success rate across the entire dataset, it still struggles with irregularly shaped objects, particularly thin and slender ones like knives and pens. Even when trained from scratch, the experts fail to perform well on these objects, indicating a need for specific design considerations.

\section{Conclusion}

This work focuses on general dexterous functional pre-grasp manipulation, repositioning and reorienting various objects to precisely match diverse functional grasp poses, crucial for real-world functional grasping. We adopt a teacher-student learning framework, introducing a novel mutual reward that greatly enhances teacher policy learning. Furthermore, we propose employing a \ac{moe} and distillation with a diffusion policy for learning diverse manipulation behavior. Our experiments showcase efficacy and robustness, revealing the potential for real-world applications.

\paragraph*{Limitations and Future Works}

Although our teacher policy shows promising results, it still struggles with objects of irregular shapes. Integrating human demonstrations could potentially improve performance. Additionally, our current focus is solely on pre-grasp manipulation. To achieve functional grasping in real-world scenarios, it is essential to integrate pre-grasp manipulation with functional grasp pose generation and grasping, alongside addressing the sim2real gap. 

\paragraph*{Acknowledgments}

We thank Jiyao Zhang, Haoran Geng, Zeyuan Chen, Tianyu Wang, Xiangyu Huang, and Jinghui Zhuang for their insightful discussions. 

\setstretch{0.905}
\bibliographystyle{IEEEtran}
\bibliography{reference_header,reference}
\end{document}